 \documentclass[pmlr,twocolumn,10pt]{jmlr} % W&CP article

\usepackage{booktabs}
\usepackage[load-configurations=version-1]{siunitx} % newer version 
\newcommand{\equal}[1]{{\hypersetup{linkcolor=black}\thanks{#1}}}
\jmlryear{2021}
\jmlrworkshop{Machine Learning for Health (ML4H) 2021} % W&CP title

\title{Explaining medical AI performance disparities across sites with confounder Shapley value analysis}

\author{%
\Name{Eric Wu}\equal{These authors contributed equally} \Email{wue@stanford.edu}\\
\addr Department of Electrical Engineering, Stanford University, USA
\AND
\Name{Kevin Wu}\footnotemark[1] \Email{kevinywu@stanford.edu}\\
\addr Department of Biomedical Informatics, Stanford University, USA
\AND
\Name{James Zou} \Email{jamesz@stanford.edu}\\
\addr Department of Biomedical Data Science, Stanford University, USA
}

\begin{document}

\maketitle

\section{Abstract}

\; Medical AI algorithms can often experience degraded performance when evaluated on previously unseen sites. Addressing cross-site performance disparities is key to ensuring that AI is equitable and effective when deployed on diverse patient populations. Multi-site evaluations are key to diagnosing such disparities as they can test algorithms across a broader range of potential biases such as patient demographics, equipment types, and technical parameters. However, such tests do not explain \textit{why} the model performs worse. Our framework provides a method for quantifying the marginal and cumulative effect of each type of bias on the overall performance difference when a model is evaluated on external data. We demonstrate its usefulness in a case study of a deep learning model trained to detect the presence of pneumothorax, where our framework can help explain up to 60\% of the discrepancy in performance across different sites with known biases like disease comorbidities and imaging parameters. 

\section{Introduction}

\; Artificial intelligence algorithms used in healthcare settings are prone to performing worse when deployed on unseen patient populations \citep{thambawita2020extensive, chirra2018empirical, song2020cross, wu2021medical}. Multi-site evaluation, where algorithms are validated on data collected from multiple sites not used in training, is an important tool for ensuring generalizability in real-world settings. Models can easily learn hospital-specific image artifacts and medical equipment settings \citep{antun2020instabilities, willemink2020preparing} or inherit biases in patient race, sex, and socioeconomic status \citep{seyyed2020chexclusion, banerjee2021reading, larrazabal2020gender}. Such variations can perpetuate disparate outcomes and mask patient harm. However, it is often unclear the degree to which each of these biases contribute to cross-site performance disparities observed in practice.
\paragraph{
}Knowledge of why an algorithm is biased can enable developers to more efficiently rectify these issues and inform its users of its limitations. In this paper, we propose a framework for explaining performance disparities in multi-site evaluations as the sum of marginal contributions from confounding site factors. We conduct a case study on pneumothorax detection models and show that potential biases such as demographic composition and image-related factors can account for up to 60\% of observed cross-site performance disparities, suggesting its usefulness as a tool for model developers and regulators to mitigate risks associated with algorithmic bias.
\paragraph{}
\; Previous work has focused on cataloging various types of machine learning bias \citep{mehrabi2021survey, gu2019understanding, char2018implementing, hitti2019proposed, gianfrancesco2018potential}. While these studies have analyzed the effect of individual bias types, our work provides a first-take at quantifying the marginal and cumulative contributions that multiple bias types may play in cross-site performance disparities. 

\begin{figure*}[t!]
\label{bars}
\floatconts
  {fig:bars}
  {\caption{Each column represents the total performance disparity between a within-site test dataset and an external dataset. Column names refer to a model trained on the \textit{first dataset} and evaluated on the \textit{second dataset}. For instance, in the "NIH on SHC" column, the bottom number refers to the test AUC of a model trained on NIH and evaluated on SHC, the middle number refers to the test AUC on SHC after matching the distribution of SHC, and the top number refers to the test AUC of the model trained and evaluated on NIH.}}
  {\includegraphics[width=0.8\linewidth]{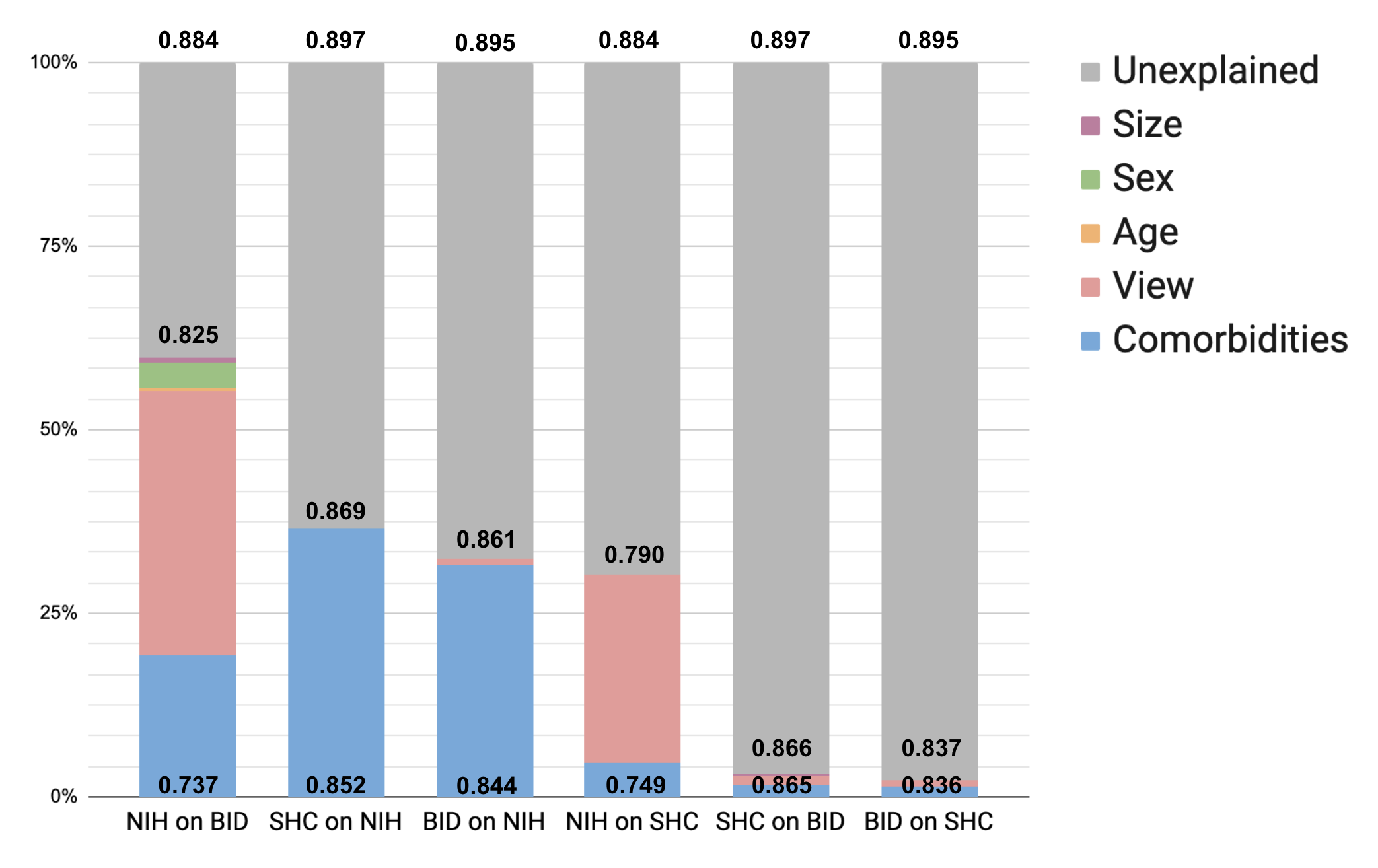}}
\end{figure*}

\section{A confounder Shapley value analysis framework}
\paragraph{}
A model's observed performance disparity between two sites can be explained as the cumulative effect of differences in the underlying patient populations and clinical standards of the two sites. For instance, one site might have older patients, which may be more challenging to an algorithm trained on younger patients. Alternatively, if one site uses imaging equipment from a different manufacturer than another, an algorithm may learn image statistics produced by one and perform poorly on images produced by the other. In practice, algorithms may have biased performance due to any number of idiosyncrasies in the patient or image acquisition composition of a clinical site.\\

Let $\mathcal{Z}=\mathcal{X}\times \mathcal{Y}$, where $\mathcal{X}$ is the data (eg. medical images) and $\mathcal{Y}$ is the associated label (eg. presence of pneumothorax). 

\begin{definition}[Site Performance]
Given a performance metric $U$, let $h^*(z)$ be a model trained on data from site $T$ such that $h^*(z) = argmin_{h\in \mathcal{H}}\mathbb{E}_{z\sim f_Z(z|d=T)}[U(z|h(z)]$.
The performance of $h^*(z)$ on data drawn i.i.d. from any distribution $f_Z$ is
$$
\Lambda(f_Z) \overset{\Delta}{=} \mathbb{E}_{z\sim f_Z}[U(z|h^*(z))],
$$

The performance disparity between sites $A$ and $B$ is $\Lambda(f_Z(z|s=A)) - \Lambda(f_Z(z|s=B))$.
\end{definition}

Let $V = \{V_1, ..., V_K\}$ be measurable random variable confounders ("site factors") that differentiate sites from one another. Each site factor $V_i$ has site-conditional distribution $f_{V_i}(v_i | s=m)$. \\
The cross-site performance disparity between sites $A$ and $B$ can be expressed as the sum of marginal contributions of each site factor, $\{\phi_1, ..., \phi_K\}$:
$$\Lambda(f_Z(z|s=A)) - \Lambda(f_Z(z|s=B)) = \sum_{i}^K \phi_i + \epsilon,$$ where $\epsilon$ is unexplained disparity due to unmeasured site factors.

The individual contribution of each site factor $V_i$ to the performance disparity is the effect of controlling for the distribution of $(V_i,f_{V_i})$ between sites.
\paragraph{}
In order to compute $\phi_i$ corresponding to each $V_i$, we use Shapley values, where the utility function is the marginal contribution of site factors. Shapley values provide a principled framework for determining individual contributions of players (or site factors) on overall utility (site performance) \citep{ghorbani2019,ghorbani2020}. In particular, the \textit{efficiency} property provides the ability to decompose the aggregate effect of site differences into constituent parts. The \textit{null} property allows regulators to determine cases where a certain site feature has no contribution to performance disparity, especially for protected factors like sex and race. The \textit{symmetry} property allows for discovery of factors which contribute equally to disparate site performances.

\begin{definition}[Cross Site Disparity]
The disparity attributable to $V_i$ is the difference in performance when controlling for $f_{V_i}$. Let $\Delta$ be a function that maps a set of site factors $\{V_1, ..., V_i\}$ to the  performance disparity between sites $A$ and $B$ after controlling for those factors.
\begin{align*}
    \Delta(\{V_i\}) &\overset{\Delta}{=}  \\
                &\Lambda(f_Z(z|d=A, f_{V_i}=f_{V_i}(v_i|s=B))) - \\
                &\Lambda(f_Z(z|d=A))
\end{align*}
\end{definition}

If $\Delta(\{V_i\})=\Lambda(f_Z(z|s=A)) - \Lambda(f_Z(z|s=B))$, then $V_i$ explains all the cross-site disparity between sites $A$ and $B$. Given an empty set, $\Delta(\emptyset) = \Lambda(f_Z(z|s=A)) - \Lambda(f_Z(z|s=B))$ is the empirical cross-site disparity.

\begin{definition}[Shapley Value Of Site Factors]
The Shapley value of $V_i$ is its contribution to cross-site disparity:
\begin{align*}
    \phi_i(\Delta) &= \\
    &\sum_{S\subseteq V\setminus\{i\}}\frac{|S|!(|V|-|S|-1)!}{|V|!}(\Delta(S\cup\{i\}) -  \Delta(S))
\end{align*}
\end{definition}

\paragraph{}
The full algorithm is described in Appendix Algorithm 1. Given a subset of site factors $S \subseteq V$ and sites $A$ (the reference site) and $B$ (the external site), we sample $Z\sim f_Z(z|S=B)$ such that $f_{V_i} = f_{V_i}(v_i|s=B)$ for all $V_i \in S$. In the case where we are controlling for multiple site factors, we match the distributions sequentially for each factor, as opposed to matching the joint distribution. As an example of our matching process, given patient race, if site $A$ contains 60\% Black/40\% White, then we resample data from site $B$ to match this proportion. The marginal performance change observed for each feature is computed and stored. Only subsets of site factors that have a sufficient number of samples to determine performance are included in the Monte Carlo simulation. We sample from S until the standard error of the Monte Carlo mean falls below a threshold (set at 0.005 in our experiments) or reaches a maximum number of iterations and report the Shapley values for each feature as the average for each feature over all samples.

\section{Case Study: Pneumothorax}
\paragraph{}
We apply this framework to a case study on detecting the presence of pneumothorax in X-ray images. We chose three publicly available datasets that contain pneumothorax as a disease outcome, sourced from different hospital sites in the USA: the National Institutes of Health Clinical Center in Bethesda, Maryland (NIH) \citep{wang2017chestx}, Stanford Health Care in Palo Alto, California (SHC) \citep{irvin2019chexpert}, and Beth Israel Deaconess Medical Center in Boston, Massachusetts (BID) \citep{johnson2019mimic}. We then trained three separate DenseNet-121 \citep{huang2017densely} deep learning models on each of the three datasets (taking as input a chest X-ray image and outputting a probability prediction for the presence of pneumothorax), and then evaluated the models on the other two datasets. The test AUC results are shown in Appendix Table 1.\\

As each dataset includes a distinct set of associated metadata (eg. patient demographic information, image acquisition parameters, diagnoses), we identified a common set of metadata present across the three datasets. This feature set includes age, sex, X-ray image view (lateral/LL, anterior-posterior/AP, or posterior-anterior/PA), X-ray image size (ratio of height to width), comorbidities (atelectasis, cardiomegaly, consolidation, edema, effusion, lesion, pneumothorax, or no finding). For each of the six cross-site evaluations performed, we report the fraction of performance discrepancy explained by each of these factors (Figure 1). \\

Overall, the framework was able to explain an average of 27\% of the total performance disparity. On two of the evaluations -- SHC on BID and BID on SHC -- the framework could only explain a negligible amount of discrepancy. As there is only limited available metadata common across the two datasets, additional factors are needed to account for the unexplained differences. \\

We observe that image view and comorbidities are the two largest contributors to performance disparity. Image views contain clinically meaningful differences in disease presentation but are often not reported or trained separately in deep learning models. Comorbidities can also present a challenge in diagnosing pneumothorax, as other conditions co-present in an image may be mistakenly flagged as positive pneumothorax cases. Of the comorbidities studied, atelectasis (collapsed lung) and cardiomegaly (enlarged heart) are the two most significant contributors to performance drop-offs. 

\section{Discussion}
\paragraph{}
A general framework for explaining sources of bias and performance disparities can (1) allow clinicians and users to understand why an algorithm performs worse on external datasets, (2) encourage algorithm developers to collect training data factors that have been demonstrated to explain the most performance drop, and (3) inform dataset curators on the types of metadata that are most useful to collect and release.
\paragraph{}
While multi-site evaluation is inherently useful, sourcing multiple high-quality external datasets pose a challenge to most algorithm developers. Reasons include the regulatory burden of releasing patient medical records from hospitals, the logistical and technical hurdle of collecting thousands of patient cases from electronic health record systems, and the high cost of sourcing reliable clinician annotations. Indeed, even among US FDA-approved medical AI algorithms, less than a third reported conducting multi-site evaluations in its approvals \citep{wu2021medical}. Furthermore, even when multi-site evaluations are conducted, the diversity of patient cases selected across the external sites can often still look very similar in terms of geographic composition \citep{kaushal2020geographic}, medical equipment used, image acquisition parameters, and other factors. In these cases, a “well-performing” multi-site evaluation can actually mask potential performance disparities and provide a misleading conclusion of generalizability.
\paragraph{}
Thus, there is a strong incentive for data-driven and more precise methods of data sourcing. By being able to target specific known factors responsible for performance drops, algorithm developers are able to focus limited resources on collecting types of data where disparities are more likely to occur. Similarly, regulators evaluating algorithms can use a data-guided approach in steering applicant candidates toward better dataset curation and evaluation.
\paragraph{}
One important limitation of our framework is that metadata associated with datasets is often unavailable, sparse, or unstructured, resulting in a limited understanding of the diversity of data available. For instance, in our case study, the performance differences between the SHC and BID datasets are largely unexplained by our method as a consequence of lacking shared metadata. Indeed, metadata such as race, socioeconomic status, and image acquisition parameters are not available in both datasets and may contain hidden sources of bias that are not yet able to be explored. High-quality metadata collection and reporting in publicly available medical datasets are thus very important for a comprehensive understanding of sources of performance differences. 
\paragraph{}
Additionally, noise present in datasets can often be quite high and therefore difficult or impossible to capture with matching distributions alone. For instance, each dataset has a distinct process for sourcing ground truth labels -- whether using single clinician vs committee decision annotations, or different NLP algorithms for parsing electronic health records. Anecdotally, this noise might have affected up to 15\% of the labels in BID dataset \citep{Oakden-Rayner2019}. Moreover, patient demographics are often self-reported or missing, and comorbidities rely on fuzzy matching from free text in clinician notes. All of these factors can make reliable explanations more difficult.
\paragraph{}
Despite these limitations, our case study provides a proof-of-concept validation of our proposed framework. While further experimentation is needed to verify its usefulness across other modalities and datasets, we believe that a principled approach toward understanding the constituent components of multi-site bias is key to more equitable model development in medical AI.

\bibliography{jmlr-sample}

\appendix

\begin{figure*}[t!]
\label{flow}
\floatconts
  {fig:flow}
  {\caption{Bias may arise from all stages of an algorithm’s lifecycle; from the types of patients included in the training dataset, the configurations used in acquiring and storing data, the parameters set when training the model, to its actual usage in a clinical setting. Properly decomposing performance disparities into individual factors like the ones listed above are important for understanding and addressing algorithmic bias in practice.}}
  {\includegraphics[width=0.9\linewidth]{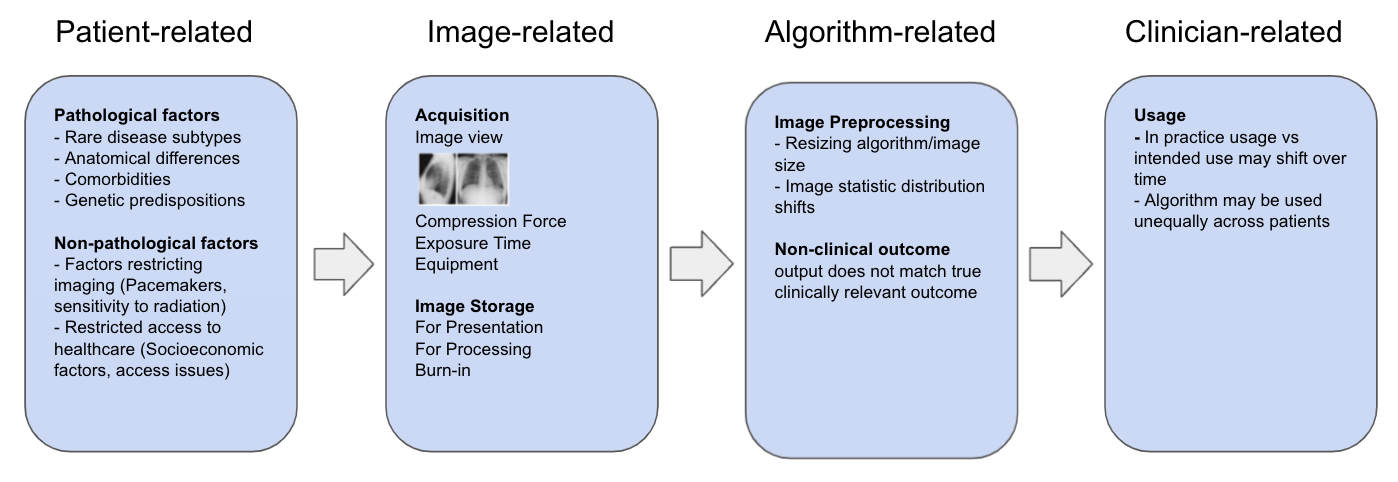}}
\end{figure*}

\begin{figure*}[t!]
\label{dist}
\floatconts
  {fig:bars}
  {\caption{Illustration of the distribution matching framework. Orange represents a model trained on data from one site and evaluated on hold-out test data from the same site. Solid blue represents that model’s performance on an external, unseen dataset. The model performs worse because of significant differences in the underlying distribution of data. After resampling the external dataset to match the distribution of the hold-out test set, the performance gap narrows significantly, as shown in striped blue.}}
  {\includegraphics[width=1.0\linewidth]{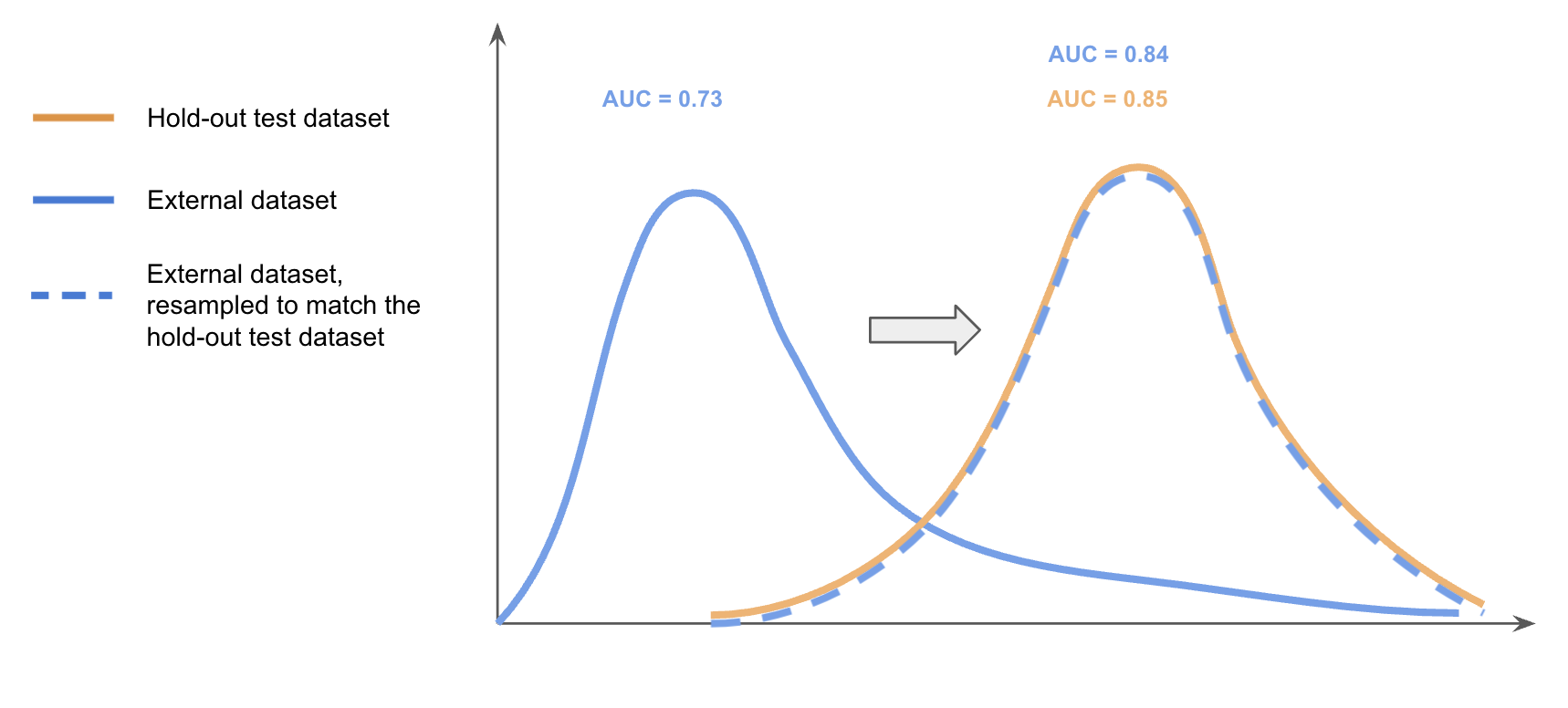}}
\end{figure*}

\begin{algorithm2e*}[t!]
\label{algo}
  {\caption{Shapley value-based distribution matching algorithm}}%

\KwIn{$V$, a list of factors; $D_0$, the test dataset; $D_1$, the external dataset; $\Lambda(D_i)$, the performance observed on $D_i$ $t$, a tolerance bound for the standard error (SE) of the Monte Carlo mean};
\KwOut{A list of Shapley values for each factor in $V$}
$N \gets length(V)$ \\
$S \gets []$ \\

\While{max iterations not reached}{
    $k \gets$ shuffle([$1$...$N$])\;
    $s \gets [\mathbf{0}_{\times N}]$\;
    $p \gets $[$\Lambda(D_0)$]\;
	\For{$i \gets 1$ to $N$}{
        $R_{i} \gets$ sample with replacement $\vert D_1\vert$ elements from $D_1$ matching the distribution of first $i$ factors of $V[k]$ in $D_0$\;
        $p \gets p \cup \Lambda(R_{i}$)\;
    }
	$s$[$k$] = [$p$[j] - $p$[j-1] for $j \gets 1$ to $N$]\;
	\If{SE($s$) $<$ t}{
        exit loop\;
    }
    $S \leftarrow s$\;
}
\Return column-wise mean of $S$ 
\end{algorithm2e*}

\begin{table*}[p]
\label{table}
\centering
\begin{tabular}{p{0.03\textwidth}lll}
\toprule
 & \bfseries BID & \bfseries SHC  & \bfseries NIH  \\
  \midrule
\bfseries BID & $0.895\pm 0.008$    & $0.836\pm 0.009$   & $0.844\pm 0.019$    \\ 
\bfseries SHC & $0.865\pm 0.010$    & $0.897\pm 0.008$    & $0.852\pm 0.018$    \\ 
\bfseries NIH & $0.737\pm 0.016$    & $0.749\pm 0.011$    & $0.884\pm 0.016$    \\ 
\end{tabular}
\caption{Cross-site evaluations of a deep learning model trained to detect pneumothorax X-ray cases. The 95\% confidence intervals are included, calculated via bootstrapping. Each cell refers to the test AUC of the row-specified model evaluated on the column-specified dataset. For example, the (BID, BID) entry is the model trained on BID and tested on held-out BID patients. The (BID, SHC) entry is the BID model evaluated on SHC test patients.}
\end{table*}

\begin{table*}[p]
\centering

\label{table}
\begin{tabular}{p{0.15\textwidth}lllllll}
\toprule
  \bfseries Evaluation & \bfseries Comorbidities & \bfseries View  & \bfseries Age   & \bfseries Sex   & \bfseries Size   & \bfseries Unexplained & \bfseries Total\\
  \midrule
\bfseries NIH on SHC & 0.006         & 0.034 & 0.000 & 0.000 & 0.000 & 0.094       & 0.135            \\
\bfseries NIH on BID & 0.028         & 0.053 & 0.001 & 0.005 & 0.001 & 0.059       & 0.147            \\
\bfseries SHC on NIH & 0.016         & 0.000 & 0.000 & 0.000 & 0.000 & 0.029       & 0.045            \\
\bfseries SHC on BID & 0.001         & 0.000 & 0.000 & 0.000 & 0.000 & 0.031       & 0.032            \\
\bfseries BID on NIH & 0.016         & 0.000 & 0.000 & 0.000 & 0.000 & 0.034       & 0.051            \\
\bfseries BID on SHC & 0.001         & 0.000 & 0.000 & 0.000 & 0.000 & 0.058       & 0.059           
\end{tabular}
 
\caption{{The marginal contributions from each of the six metadata factors. Each number refers to the marginal difference in AUC explained by a feature between a within-site test AUC and the external evaluation AUC.}}
\end{table*}

\end{document}